\documentclass[sigconf]{aamas} 

\usepackage{balance} 
\usepackage
{subcaption} 
\usepackage{algorithm}
\usepackage{algpseudocode}

\acmSubmissionID{699}

\title[AAMAS-2023 Formatting Instructions]{AI Agent as Urban Planner: Steering Stakeholder Dynamics in Urban Planning via Consensus-based Multi-Agent Reinforcement Learning}

\author{Kejiang Qian}
\affiliation{
  \institution{King's College London}
  \city{London}
  \country{United Kingdom}}
\email{kejiang.qian@kcl.ac.uk}

\author{Lingjun Mao}
\affiliation{
  \institution{Tongji Univeristy}
  \city{Shanghai}
  \country{China}}
\email{mao1207@tongji.edu.cn}

\author{Xin Liang}
\affiliation{
  \institution{Tongji Univeristy}
  \city{Shanghai}
  \country{China}}
\email{2053246@tongji.edu.cn}

\author{Yimin Ding}
\affiliation{
  \institution{Tongji Univeristy}
  \city{Shanghai}
  \country{China}}
\email{2153487@tongji.edu.cn}

\author{Jin Gao}
\affiliation{
  \institution{Massachusetts Institute of Technology}
  \city{Cambridge}
  \country{USA}}
\email{gaojin@mit.edu}

\author{Xinran Wei}
\affiliation{
  \institution{Tongji University}
  \city{Shanghai}
  \country{China}}
\email{xinran_wei@tongji.edu.cn}

\author{Ziyi Guo}
\affiliation{
  \institution{University of Pennsylvania}
  \city{Philadelphia}
  \country{USA}}
\email{guoziyi@upenn.edu}

\author{Jiajie Li}
\affiliation{
  \institution{Massachusetts Institute of Technology}
  \city{Cambridge}
  \country{USA}}
\email{jiajie@mit.edu}

\begin{abstract}
In urban planning, land use readjustment plays a pivotal role in aligning land use configurations with the current demands for sustainable urban development. However, present-day urban planning practices face two main issues. Firstly, land use decisions are predominantly dependent on human experts. Besides, while resident engagement in urban planning can promote urban sustainability and livability, it is challenging to reconcile the diverse interests of stakeholders. To address these challenges, we introduce a Consensus-based Multi-Agent Reinforcement Learning framework for real-world land use readjustment. This framework serves participatory urban planning, allowing diverse intelligent agents as stakeholder representatives to vote for preferred land use types. Within this framework, we propose a novel consensus mechanism in reward design to optimize land utilization through collective decision making. To abstract the structure of the complex urban system, the geographic information of cities is transformed into a spatial graph structure and then processed by graph neural networks. Comprehensive experiments on both traditional top-down planning and participatory planning methods from real-world communities indicate that our computational framework enhances global benefits and accommodates diverse interests, leading to improved satisfaction across different demographic groups. By integrating Multi-Agent Reinforcement Learning, our framework ensures that participatory urban planning decisions are more dynamic and adaptive to evolving community needs and provides a robust platform for automating complex real-world urban planning processes.

\end{abstract}


\newcommand{\BibTeX}{\rm B\kern-.05em{\sc i\kern-.025em b}\kern-.08em\TeX}

\begin{document}


\pagestyle{fancy}
\fancyhead{}


\maketitle 


\section{Introduction} 

As cities develop and transform, previous land-use configurations may become outdated and inadequate for current needs. Urban planner provides an approach of land use readjustment to optimize land use distribution, aligning the urban environment with evolving challenges arising from population growth, aging infrastructure, and environmental considerations\cite{yang20201}. Its core processes involve identifying existing land use issues, suggesting potential solutions, and implementing the most viable ones. 

Conventional urban planning is typically an experience-driven process that follows a top-down approach, executed by governments, urban planners, and developers~\cite{taylor1998urban}. However, this approach often struggles to advance urban readjustment plans, faces legal conflicts due to compulsory acquisitions, or encounters disagreements from individuals~\cite{hong2012land}. Furthermore, it is challenging to reconcile the land use preferences of various stakeholders. A participatory urban planning strategy has been put forward \cite{laurian20091}, aimed at bolstering public participation in decision-making, reconnecting connections between citizens and decision-makers, and fostering consensus in urban planning. Nonetheless, issues have arisen concerning the difficulty of enhancing their trust and the occasional emergence of unforeseen outcomes \cite{astrom20201}.

Existing research has been utilizing computational approaches to tackle the trust crisis and foster collective decision-making in land use readjustment. Among the methods explored, agent-based modeling~\cite{parker2003multi} stands out as a prominent method. Initially, it simulates the behaviors of diverse stakeholders, represented as agents, based on predefined rules derived from expert experience. Subsequently, it captures the nexus of public and private interests in this decision-making and projects potential land use changes~\cite{GROENEVELD201739,ghavami2017towards}. However, there are three key challenges that limit its adaptability in real-world applications. Specifically, the foremost challenge lies in its heavy reliance on predefined rules and expert knowledge, and cannot adapt to the evolving urban environments. Secondly, cities operate as systems, with land use changes being deeply interconnected rather than isolated events \cite{batty20131}. The third challenge arises from the complexity of modeling stakeholder interactions, especially when considering individual land use preferences. It is exacerbated by the cascading chain reactions resulting from the intertwined nature of urban dynamics and stakeholder dynamics.

Addressing this challenge, we introduce a consensus-based multi-agent reinforcement learning (MARL) algorithm for urban planning. This algorithm aims to enhance urban benefits while recognizing and catering to the interests of each individual agent. Unlike MARL applications in some simpler urban scenarios, we adopt a graph-based data structure that encapsulates the geographical information of the urban area. In this structure, land parcels are represented as nodes, while edges connect adjacent parcels. To effectively process its spatial features, Graph Neural Networks (GNNs) are integrated into the Actor and Critic networks designated for the agent's learning. Distinctly, agents are classified into two primary groups: the top-down and the bottom-up. The top-down group contains entities like urban planners and developers, while the bottom-up group subdivides into representatives from the lower, middle, and upper socioeconomic brackets. To steer stakeholder dynamics, we propose four levels of cooperative metrics for reward and learning objective design to build consensus among agents and reach an optimal land use readjustment strategy. Furthermore, we apply it to real-world land use readjustments in the chosen study area, Kendall Square, to assess the performance of our framework. Extensive experiments indicate that our framework produces optimal planning outcomes, marking a notable enhancement over traditional computational methods.





\section{Related Work} 



Land use readjustment stands as a critical topic in spatial optimization research \cite{WANG2021107540,LIAO2022108710}. Conventional techniques such as multi-objective optimization \cite{LAN20157334792} and swarm intelligence algorithms \cite{GAO2017978} have been leveraged to allocate land resources, aiming for diverse objectives ranging from maximizing economic output to achieving sustainability goals. Besides, Abolhasani et al. \cite{ABOLHASANI2022101795} emphasized the importance of collective decision-making and stakeholder interdependencies in urban planning. However, these methods often are rigid and lack the flexibility to adapt to changing urban scenarios and balance stakeholder interests.

In response to the challenges posed by conventional methods, deep reinforcement learning (deep RL) has emerged as a powerful approach for optimizing nonlinear strategies. The advances in deep RL have found applications across various domains, including robotics system \cite{gu2017deep}, gaming \cite{berner2019dota}, and protein structure prediction \cite{jumper2021highly}. In urban computing, deep RL is widely used in operation problems of different urban systems. Liang et al. \cite{Liang20198600382} introduced a deep reinforcement learning model that utilizes real-time traffic data from sensors to optimize traffic light duration, significantly reducing average waiting times and outperforming traditional methods in both rush hour and regular traffic conditions. Xi et al. \cite{Xi20229839295} developed a hierarchical mixed deep reinforcement learning approach for efficiently repositioning vehicles in online ride-hailing systems, demonstrating superior performance in simulations using real order data, with plans for future enhancements, including graph-structured maps and advanced prediction algorithms. Li et al. \cite{Li20189781} proposed a spatio-temporal reinforcement learning approach for optimizing bike repositioning in bike-sharing systems, utilizing a unique clustering algorithm to reduce complexity, and demonstrating its effectiveness through real-world Citi Bike data experiments. However, these studies focus on relatively simpler environments than urban planning problems without complicated agent interactions and it is less feasible to directly utilize their methods to this challenge.

Besides, recent advanced research focuses on enhancing decision-making of urban interventions for experts, such as urban planning, and economic policy. The AI Economist \cite{Stephan20221} is a novel deep RL framework that co-adapts agents and a social planner for economic policy design, demonstrating improved outcomes in taxation. Although it has not been calibrated with real-world settings, it provides a test bed for policy-making. Zheng et al. \cite{zheng20231} developed an AI-based urban planning model that outperforms human experts by using graph neural networks to generate efficient spatial plans, suggesting a collaborative approach for future urban development. These papers provide evidence that deep RL can enhance the decision-making of complex intervention strategies.

Urban planning is a complicated process containing dynamic interactions among various stakeholders, including the government, urban planners, developers, and citizens, rather than just selecting land use sites \cite{zheng20231}. A central challenge is to balance the individual preferences of these agents, who have diverse land use priorities, with the collective interests of the broader urban landscape. While traditional MARL works \cite{hua20231,Guresti20231,figura20211} have made significant advancements by addressing both individual and collective agent interests, its application in urban planning remains a nascent area of exploration. Despite some methods achieving performance improvements in specific contexts, there is a need to delve deeper into the potential implications of MARL in urban settings, particularly concerning the understanding of its impacts on urban systems and the design of reward structures for stakeholders.




\section{PROBLEM FORMULATION} 
In this section, we define the urban readjustment problem using a spatial graph, denoted as \(G=(P, E, N)\), where \(P\) is  the foundational urban parcels, \(E\) indicates the edges connecting these parcels, and \(N\) is the representatives of stakeholder groups in the urban landscape. More specifically,


\begin{itemize}
    \item Each parcel \(P_i \in P\) stands for a region that contains its land features \(P=(L \cup A)\) to capture urban geographical layout, in which \(L\) and \(A\) stand for the set of parcel land uses and their respective areas. Specifically, its land use type \(L\) is categorized into five general land use types: residential zone \(r\), office \(o\), green space \(g\), commercial zone \(c\), facilities \(f\), denoted by \(L=\{r, o, g, c, f\}\). 
    \item Each edge \(E_i \in E\)  in the graph represents the spatial relationships and the Euclidean distance \(d\) between a pair of parcels. It is denoted by an adjacency matrix \(X_{ij}\) where,
$$X_{ij}= \begin{cases}
d & if\ there\ is\ an\ edge\ between\ P_i\ and\ P_j \\
0 & otherwise
 \end{cases}$$
An adjacency list of parcel \(P_i\) is given as \([(P_j, d_1), (P_k, d_2),...]\). This format provides a detailed perspective on their spatial proximity and interrelation.
\item

In the context of urban development, stakeholders are represented by the set \(N\). These stakeholders are broadly categorized into residents, \(N_{\text{residents}}\), and urban professionals, \(N_{\text{professionals}}\).

The residents set, \(N_{\text{residents}}\), comprises:
\[ N_{\text{residents}} = N_{\text{low}} \cup N_{\text{mid}} \cup N_{\text{high}} \]
where \(N_{\text{low}}\), \(N_{\text{mid}}\), and \(N_{\text{high}}\) represent the low, middle, and high-income bracket residents, respectively.

The urban professionals set, \(N_{\text{professionals}}\), is defined as:
\[ N_{\text{professionals}} = N_{\text{planners}} \cup N_{\text{developers}} \]
with \(N_{\text{planners}}\) representing urban planners responsible for strategizing and envisioning the urban development, and \(N_{\text{developers}}\) denoting developers who actualize these urban projects.

The overarching set of stakeholders is:
\[ N = N_{\text{residents}} \cup N_{\text{professionals}} \]
This representation highlights the varied interests and dynamics within urban development.


\end{itemize}


In terms of the MARL environment, we delve into a spatiotemporal urban readjustment simulation: participatory planning. Compared to traditional top-down planning which is a hierarchical method where decisions are made by urban planners, it is a more decentralized approach that emphasizes multiple residential agent involvement and corporations to create solutions that benefit a broader range of stakeholders and localized impacts.

Hence, introducing this environment provides urban planners with a rich test bed to test various scenarios and predict outcomes. This dynamic nature provides optimal strategies and offers a comprehensive and holistic tool for urban adjustment, ensuring a balance between macro-level development goals and micro-level community needs.

\section{Methods}
\subsection{Urban readjustment modeling}
The challenge of urban readjustment can be systematically modeled as a Markov decision process (MDP), denoted by \((N, \mathcal{S}, \mathcal{A}, \mathcal{R}, \mathcal{P}, \gamma)\). In this representation, \(N\) is the agent set, \(\mathcal{S}\) is the state set, \(\mathcal{A}\) is the action set, \(\mathcal{R}\) is the reward function, \(\mathcal{P}\) is the probability transition function, and \(\gamma\) is the discount factor. Detailed explanations of each definition are shown as follows: 

\begin{figure}[htp]
\centering
\includegraphics[width=\linewidth]{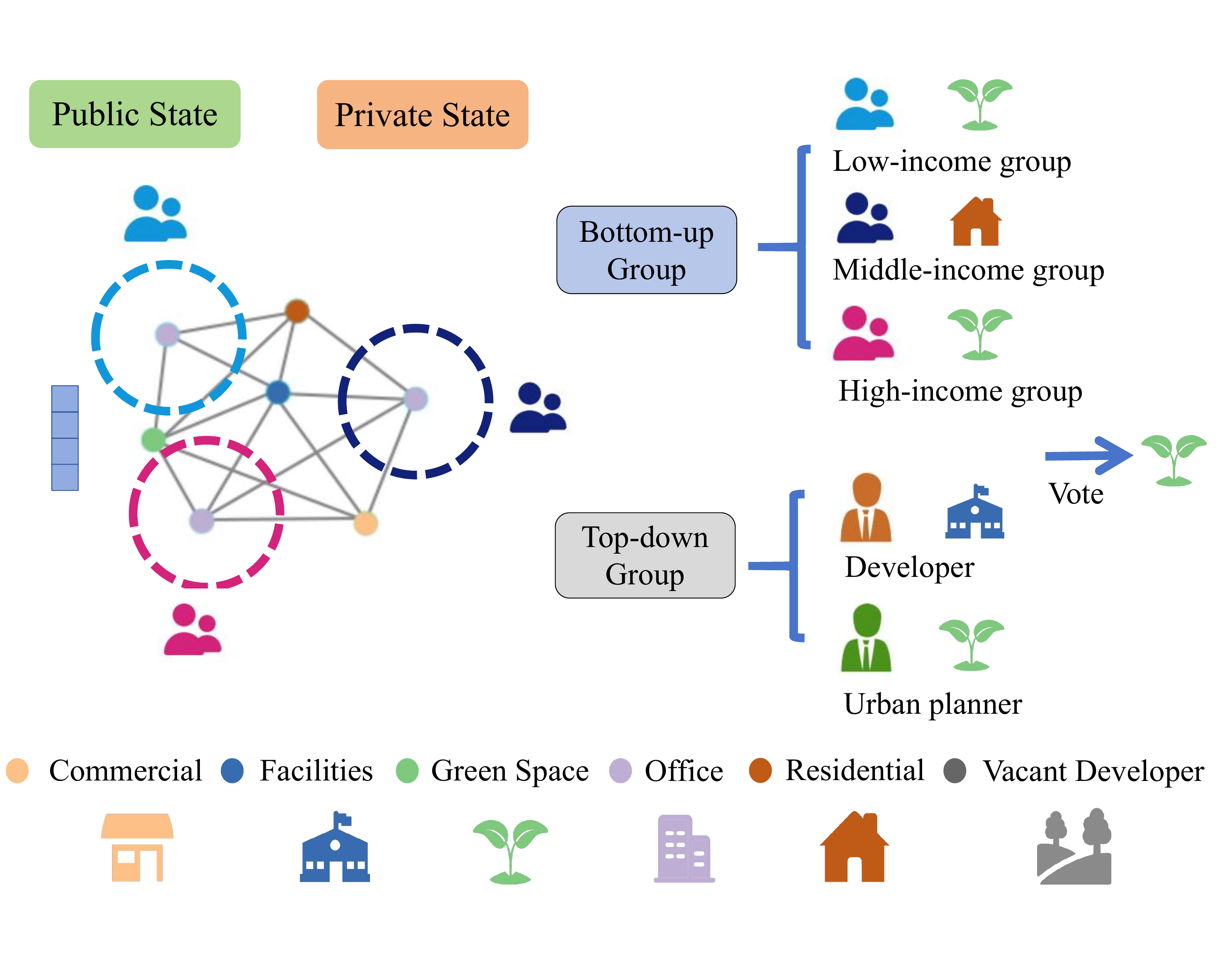}
\caption{In the Consensus-based Multi-Agent Reinforcement Learning (MARL) framework, agents are distributed across different locations with varied observation ranges. Each agent casts a vote for its preferred land use type. The collective voting outcome determines the land use type to be readjusted in the corresponding urban parcel.}
\label{fig: reward}
\end{figure}


\textbf{Agent Set} \(N\). Representatives from each of these stakeholder groups, \(N_{\text{professionals}}\)   and \(N_{\text{residents}}\)  , are defined as agents to reduce the complexity in MARL. The participatory planning environment integrates both top-down agents \(N_{\text{professionals}}\)   and bottom-up agents \(N_{\text{residents}}\), emphasizing collaboration and collective decision-making. By combining the perspectives of both groups in these two environments, urban readjustment strategies are holistic and inclusive considering the needs and priorities of stakeholders. In MARL, agents with similar characteristics, like residents of the same income bracket, can adopt a shared policy so as to significantly simplify the framework and accelerate the learning process.

\textbf{State Set} \(\mathcal{S}\). This set represents comprehensive information about the stakeholders and urban landscape. It contains details about the stakeholder's position and counts as well as all potential configurations of the urban environment, encompassing aspects such as geographical distribution, land use types, and the area of each parcel. In detail, the geographical distribution provides essential spatial information of this landscape, offering insights into the position of each parcel and its relationship to neighboring parcels. Besides, the area \(A\) of each parcel is also a dominant factor as it determines land use possibilities and development constraints.

 
\textbf{Action Set} \(\mathcal{A}\). In the Consensus-based MARL system, the overall action space, denoted as $\mathcal{A}$, represents the set of all possible action combinations of the individual agents. This global action space is defined by the formula $\mathcal{A} = \prod_{i=1}^{n} A_i$, where $A_i$ is the action space of the $i^{th}$ agent and $n$ is the total number of agents in the system.
 
This multi-dimensional action space contains all potential decisions associated with the land use type readjustment for a designated parcel \(P_j\). To be more concrete, we define \(a_i\) as the decision made by an agent \(N_i \in N\). Its feasible action space, denoted as \(A_i=[0, 1, 2, 3, 4]\). In this set, an action \(a_i \in \{0, 1, 2, 3, 4\}\) is mapped to a reassignment of a new land use type from the predefined set \(L=\{r, o, g, c, f\}\) to the parcel \(P_j\). 


\textbf{Reward Function} \(R\). The goal of urban readjustment is to balance these macro-level strategies with the micro-level needs of individual residents in a participatory planning environment. Tailored to each agent's role, it promotes decisions aligning with broader urban readjustment objectives and the agent's specific interests. For instance, a developer agent might be rewarded for profitability, while a resident agent values quality-of-life improvements. Besides, rewards can be influenced by both individual actions and the collective decisions of multiple agents. Subsequent sections will delve deeper into holistic reward functions for each agent.

\textbf{Probability Transition Function} \(\mathcal{P}\). This function captures the dynamics of the urban landscape. For a given state-action pair, it provides the probability of moving to a subsequent state, defining as a mapping \(\mathcal{S}\times\mathcal{A}\times\mathcal{S}\rightarrow[0, 1]\).


\subsection{Design of reward metrics and learning objectives}
In our proposed framework, the application of the MARL approach to real-world dilemmas requires a dedicated design of reward metrics and learning objectives, guiding the agents toward optimal decisions. This metrics design begins by assessing the depth and breadth of agents' cooperative awareness. We derive three foundational tiers of design metrics: \textit{self awareness}, \textit{local awareness}, \textit{global awareness}, and \textit{equity awareness} tailored. Each tier is to capture varying dimensions of agent interactions and their awareness levels. More specifically,
\begin{itemize}
    \item Agents with \textit{self awareness} are predominantly self-centered, focusing on their own immediate interests and land use preferences.
    \item Agents with \textit{local awareness} prioritize spatial information and the interests of their neighboring parcels when making decisions. 
    \item Agents operating with \textit{global awareness} adopt a more holistic scope, transcending their individual land use preference to prioritize broader urban performance metrics according to urban planning targets.
    \item Agents with \textit{equity awareness} are designed to recognize the diverse needs of various stakeholders, balancing these interests to reach equitable decision-making.
\end{itemize}

\subsubsection{Self Awareness.} 


In an endeavor to systematically represent the anticipated benefits or proclivities of various stakeholders towards differing land use types, we introduce a matrix representation, designated as the "Expected Benefit Scoring Matrix." This matrix \(EM \) serves as a consolidation of scores, categorized distinctly into: 0 (indicative of no expected benefit), 0.5 (representing a moderate expected benefit), and 1 (signifying a maximal expected benefit).

\[
EM = \begin{bmatrix}
1 & 1 & 1 & 0.5 & 0.5 \\
1 & 1 & 1 & 1 & 1 \\
0 & 1 & 1 & 0 & 0.5 \\
1 & 0.5 & 1 & 0.5 & 0.5 \\
1 & 0 & 0.5 & 1 & 0.5 \\
\end{bmatrix}
\]

In this matrix:
\begin{itemize}
\item The rows, from top to bottom, represent the agents \(N_{\text{planners}}\) , \(N_{\text{developers}}\), \(N_{\text{low}}\), \(N_{\text{mid}}\) and  \(N_{\text{high}}\) respectively.
\item The columns, from left to right, denote the land use types \(r\), \(o\), \(g\), \(c\), and \(f\) respectively.
\end{itemize}

Each matrix element \(EM_{ij}\) signifies the expected benefit score of agent \(i\) for land use type \(j\). For instance, the element in the third row and fourth column \(EM_{24}\) is 0, indicating that agent \(S_{b,1}\) perceives no expected benefit from the commercial zone \(c\).

This matrix representation provides a compact visualization of the relationship between agents and their respective preferences for different land use types.


\subsubsection{Local Awareness.} Local awareness refers to the agent \(N_j\) understanding of its immediate surroundings including both the physical environment and the acceptance of agents' decisions. 

A key concept encoding this awareness is the idea of the 15-minute community. This urban planning principle emphasizes the walkable accessibility of essential amenities, approximated as 1250 meters \cite{Carles20221}. In the framework, bottom-up agent \(N_{\text{residents}}\) prioritizes parcels within this range, \(dist(P_N,P_i)\), when making land use decisions, reflecting their localized interests. In contrast, top-down agents \(N_{\text{professionals}}\)   adopt a broader view of the urban landscape. 

Another aspect of \textit{local awareness} is the dynamic nature of urban readjustment. As an agent's land use type decision for parcel \(P_i\) is ratified, the intensity of their subsequent demands diminishes. This is modeled by progressively halving the reward of repeated specific requests. In summary, the rewards \(r_{L}\) can be formed by:

\begin{equation}
   r_{L}= n\cdot r_I
\end{equation}

Here, \(n\) is the number of times that preference has been considered and \(t\) presents the current round.

\subsubsection{Global awareness.} It emphasizes the objectives and targets set by urban planning initiatives to align agent decisions with overarching urban development goals.
 These requirements, derived from comprehensive studies and stakeholder consultations, form the bedrock of global awareness. To ensure that agent decisions align with these broader objectives, we introduce two primary metrics: Density and Diversity \cite{alonso20181}.

The combined reward for global awareness, \(r_G\) , is the sum of rewards derived from these two metrics:

\begin{equation}
    r_G=r_{G,1} + r_{G,2}
\end{equation}

\begin{itemize}
    \item Density \(r_{G,1}\): This metric evaluates the concentration of a specific land use type \(L\) within a district. A higher density of a required land use type, like residential zones in Kendall Square, would yield a higher reward. Mathematically, the density reward for a land use type \(m\) is given by:

\[r_{G,1}=\sum^L_{i=1}\frac{M_i}{A}\] where \(M\) is the set of occurrences of land use type \(l\) in the district and \(A\) is the area of this district.

    \item Diversity \(r_{G,2}\): Urban environments thrive on diversity, ensuring a balanced mix of land use types that cater to various needs. A district with a diverse mix of land use types would be more resilient and adaptable. The diversity reward is calculated using the Shannon Index, a commonly used metric in ecology to measure species diversity in a community:
\[r_{G,2}=-\sum^M_{i=1}p_i\cdot\ln p_i\]
where M is the total count of land use types in the district and \(p_i\) is the proportion of the occurrences of the land use type \(l_i\).
\end{itemize}

\subsubsection{Equity awareness.} The equitable distribution of decision acceptance among agents operating at different hierarchical levels. It ensures that the decisions made are not only optimal but also equitable, taking into consideration the preferences and needs of diverse stakeholders.

The hierarchical structure in the framework is twofold:
\begin{itemize}
\item Intra-group Equitability (among bottom-up group): This level focuses on the bottom-up group, which consists of agents representing different income brackets. The objective is to ensure that the decisions made are equitable among these agents, preventing any particular income group from dominating the decision-making process.

\item Inter-group Equitability (between top-down and bottom-up groups): At this level, the focus shifts to enhance the equity between citizens and decision makers in decision-making process,
balancing the decisions made by the top-down agents (e.g., urban planners and developers) with those of the bottom-up group. It ensures that the broader urban planning objectives are met without sidelining the preferences of the local community.
\end{itemize}

To quantify the equitability of decision acceptance, the reward can be presented as:
\begin{equation}
    r_{E} = -\left[\text{std}(n_{\text{low}}, n_{\text{mid}}, n_{\text{high}}) + \left| n_{\text{professionals}} - n_{\text{residents}} \right|\right]
\end{equation}
where  \(\text{std}(n_{\text{low}}, n_{\text{mid}}, n_{\text{high}})\)  computes the standard deviation of decision acceptance among the bottom-up agents representing different income brackets.



\subsubsection{Learning objectives.} In a participatory planning environment, both the top-down group agents and bottom-up group agents \(N_{\text{residents}}\) operate with these four awareness levels to ensure strategic and equitable decision-making, resonating with the broader objectives of urban readjustment. The reward function \(r\) is:
\begin{equation}
r=\beta_1 \cdot r_I+\beta_2 \cdot r_L+\beta_3 \cdot r_G+\beta_4 \cdot r_E
\end{equation}
Where \(\beta_j\)
, are hyper-parameters that dictate the significance of each awareness level.


\subsection{Model construction and training}


In the realm of reinforcement learning, the Actor-Critic framework presents a sophisticated balance between policy optimization and value estimation. The "actor" component, delineated as a neural network, is responsible for producing a probabilistic policy over actions conditioned on the observed state. This formulation ensures a degree of stochasticity, fostering both exploration of the state-action space and exploitation of known rewards. On the other hand, the "critic" component provides a value function approximation for the chosen actions in conjunction with their respective states, thereby serving as a benchmark against which the actor's decisions can be evaluated.

\begin{figure*}[htp]
\centering
\includegraphics[width=0.8\linewidth]{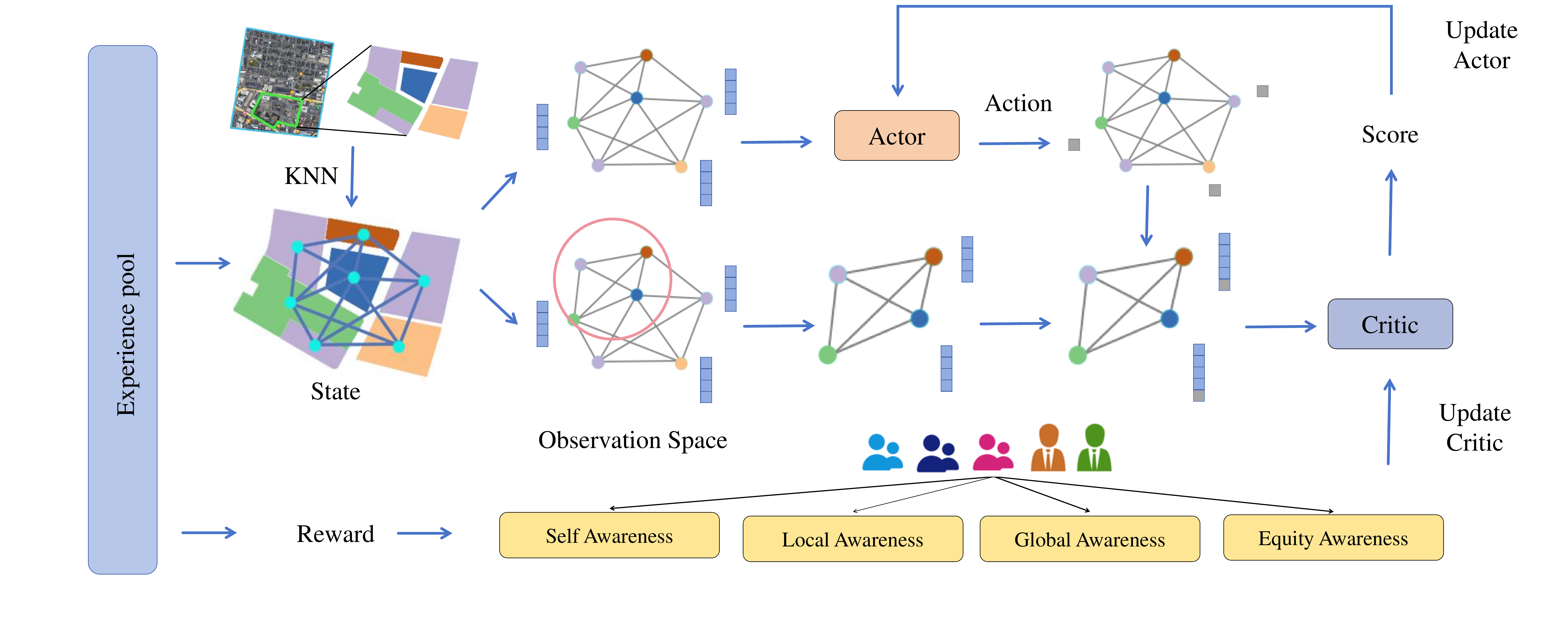}
\caption{In our framework, we utilize the Actor-Critic architecture. The Actor processes an initial graph to generate an action. Conversely, the Critic takes a combined input of the initial graph and the Actor's output, subsequently producing a score for the action in its current state.}
\label{fig:ac}
\end{figure*}

The synergy between these two networks is pivotal. Feedback from the critic facilitates gradient-based optimization of the actor's policy, nudging it towards actions that maximize expected cumulative rewards. Concurrently, the critic benefits from the trajectory samples provided by the actor, allowing for a more accurate approximation of the value function. In this treatise, we elucidate the nuanced architectural and algorithmic interplay of the Actor-Critic paradigm in Figure \ref{fig:ac}.

\textbf{Voting Policy Network(Actor)}.
The Voting Policy Network, synonymous with the Actor in Actor-Critic frameworks, is architecturally designed to generate a distribution over potential actions based on a given state. Subsequently, it suggests an action for execution in the environment. During the voting process, agents select an unvoted node and cast their vote for its intended land-use type. The Voting Policy Network receives node embeddings as input, encapsulating features along with their geographic context within the graph. To transition these node features, we apply the Graph Attention Networks (GAT) as follows:

Using the Graph Attention Networks (GAT), each node's features are first linearly transformed:
\begin{equation}
h'_{i} = W h_{i}
\end{equation}

Where \( h_{i} \) represents the feature vector of the node \( i \), and \( W \) is the weight matrix associated with the GAT layer. Attention coefficients between nodes are then computed:
\begin{equation}
e_{ij} = \text{LeakyReLU}(\mathbf{a}^T [W h_i || W h_j])
\end{equation}

Here, \( \mathbf{a} \) represents the attention mechanism's learnable weight vector. These coefficients are subsequently normalized:
\begin{equation}
\alpha_{ij} = \text{softmax}_j(e_{ij}) = \frac{\exp(e_{ij})}{\sum_{k \in \mathcal{N}(i)} \exp(e_{ik})}
\end{equation}

Finally, the new node features are determined using the attention coefficients:
\begin{equation}
h''_{i} = \sigma\left(\sum_{j \in \mathcal{N}(i)} \alpha_{ij} W h_j \right)
\end{equation}

Post-transition, the softmax function is applied to derive a policy distribution:
\begin{equation}
a_{i} = \text{softmax}(h''_{i})
\end{equation}

\textbf{Value Network(Critic)}. The Value Network, termed the Critic in Actor-Critic frameworks, is to estimate the expected return of a policy. It evaluates the potential reward for taking an action under a specific policy in its current state.

In realistic scenarios, the impact of a parcel on residents diminishes with increasing distance. Thus, we have incorporated the concept of a "15-minute living circle" into our approach. For the common resident agent, the input to the Critic isn't the entire graph. Instead, it is an observation subgraph, consisting of all parcel nodes reachable from the agent's current node within a 15-minute timeframe. Parcels beyond this observation range are deemed to have negligible influence on these agents and are thus disregarded in the analysis. In the observation subgraph, the node embeddings are constructed by concatenating the original node features and the outputs from the Actor, represented as:
\begin{equation}
\text f_i = h_i | p_{i}
\end{equation}

By integrating the Actor's input into the node embedding, we can utilize the Critic to evaluate and score the Actor's actions under the current state. After passing through multiple layers of GAT, the aggregated node features are then channeled through a fully connected layer to yield the value:
\begin{equation}
Q(s_i, a_i) = \text{FC}(\text{mean}(f_i^{(l)}))
\end{equation}



\textbf{Actor Update.}
The Actor strives to maximize the expected return as estimated by the Critic. Thus, the loss function for the Actor is the negative expected return:

\begin{equation}
\text{Loss}_{\text{actor}} = -\frac{1}{n} \sum_{i=1}^{n} Q(s_i, a_i)
\end{equation}
In this equation, $V(s_i, a_i)$ denotes the return as estimated by the Critic for the given state-action pair.

\textbf{Critic Update.}
The Critic aims to learn a value function to estimate the expected return of a given state-action pair. The loss function for the Critic is the mean squared error (MSE) between the actual returns and the returns predicted by the Critic network:
\begin{equation}
\text{Loss}_{\text{critic}} = \frac{1}{n} \sum_{i=1}^{n} (R(s_i, a_i) - V(s_i, a_i))^2
\end{equation}

Where $R(s_i, a_i)$is the actual return for the state-action pair, and $V(s_i, a_i)$ is the value predicted by the Critic for that state-action pair.




\begin{algorithm}
\caption{Consensus-based MARL Training}
\begin{algorithmic}[1]

\State \textbf{Input:} 
Graph \(G(P, E, N)\), \begin{align*}
agents \  N = \{ & N_{planners},N_{developers}, \\
& N_{low}, N_{mid}, N_{high} \}  
\end{align*}


\State Initialize experience buffers \(\text{Buffer}_{Agent}, \forall Agent \in N\)

\For {each epoch}
    \For {agent \(Agent \in N\)}
        \If {\(Agent \in \{N_{low}, N_{mid}, N_{high}\}\)}
            \State \( s_{0,Agent} \) = BFS\_subgraph(\(G\), \(Agent\), 15-min)
        \Else
            \State \( s_{0,Agent} \) = \(G\)
        \EndIf
        \State Embed \( s_{0,Agent} \) with \(\pi_{\text{Actor}_{Agent}}(s_{0,Agent})\)
    \EndFor
    
    \For {t = 1, 2, ... until done}
        \State \( a_{t} = \{\pi_{\text{Actor}_{Agent}}(s_{t-1,Agent}) | Agent \in N\} \)
        \State Execute \( a_{t} \) in \(G\), observe \( r_t, s_t \)
        \For {agent \(Agent \in A\)}
            \State \(\text{Buffer}_{Agent} \leftarrow (s_{t-1,Agent}, a_{t,Agent}, r_t, s_t)\)
            \State Sample \((s, a, r, s')\) from \(\text{Buffer}_{Agent}\)
            \State Update \(\text{Critic}_{Agent}\) using \( \nabla_{\theta_{Agent}} \mathcal{L}_{\text{critic}} \)
        \EndFor
        \For {agent \(Agent \in N\)}
            \State Compute advantage using \(\text{Critic}_{Agent}\)
            \State Update \(\text{Actor}_{Agent}\) with gradients \( \nabla_{\theta_{Agent}} \mathcal{L}_{\text{actor}} \)
        \EndFor
    \EndFor
\EndFor

\end{algorithmic}
\end{algorithm}


\section{Experimental validation-case study of Kendall Square} 
\subsection{Study area}

We selected Kendall Square, Cambridge, Massachusetts, as the study area. Historically transformed from its 1977 Major Plan Amendment, the square shows the intersection of advanced technology and business, with a dense presence of high-tech enterprises, biotech entities, and emerging startups. Its residents are distinguished by their high educational achievements and forward-thinking mindset. However, this vibrancy contrasts sharply with the northern part of the square, which features low-density housing, utilities like power plants, and a significant vacancy rate. Our goal is to offer an adaptive land use readjustment solution for this area, testing the feasibility of our proposed framework in a real-world setting..

\subsection{Data processing}
The geographical data of Kendall Square is collected from \href{https://www.cambridgema.gov/gis}{the open-source database of the City of Cambridge}. The dataset comprises 749 parcels, each characterized by land use type (categorical), parcel area (numerical), and geographic coordinates (latitude and longitude). There are three steps to process this dataset, including parcel clustering, selecting readjustment parcels, and utilizing the dataset. More specifically,

\textit{Parcel clustering.}
To model the land use distributions in Kendall Square as a spatial graph, an adjacency matrix \(X_{ij}\) is constructed using kernel-based spatial weights along with the \(k\) nearest neighbors (KNN) approach, setting \(k=4\) for every parcel.

\textit{Selecting readjustment parcels.}
We selected the readjustment parcels based on the following criteria: (1)
Obsolete facilities in need of renewal or facing environmental challenges, including parking buildings, factories, and the cogeneration station;
(2) Open spaces that are cost-effective for redevelopment, such as open plots and idle land;
(3) Vacant spaces, including the parcels registered as "vacant" for residential or commercial, and real estate market leasing and selling properties available on site.

\textit{Utilizing the dataset.}
During the experimentation phase, agents collaboratively decide on the land use type for readjustment parcels. To evaluate the superiority of the participatory urban planning framework, we contrasted it with traditional top-down planning in which decisions are exclusively made by an urban planner.

\subsection{Evaluation metrics} 

\subsubsection{Compared methods}
In our study which is conducted under top-down planning and participatory planning environments, we select four methods as baselines for comparison: the random method, the greedy method, DRL and our proposed method. Specifically:
\begin{itemize}
\item The random method, \underline{R}andom \underline{T}op-down \underline{P}lanning (RTP) and random participatory planning (RPP), allows each agent to vote for a land-use type based on random selection.

\item The greedy method, \underline{G}reedy \underline{T}op-down \underline{P}lanning (GTP) and \underline{G}reedy \underline{P}articipatory \underline{P}lanning (GPP) enables agents to vote for the land-use type that is most beneficial to them individually, neglecting the global impact.
\item DRL, \underline{D}RL \underline{T}op-down \underline{P}lanning (DTP) is employed in top-down planning, centralizing decision-making of land readjustment.
\end{itemize}

\subsubsection{Algorithmic metrics}
To assess the performance of compared methods from an algorithmic standpoint, we focused on two key rewards, Global Reward and Equity Reward, derived from the reward metric=s aforementioned: global awareness and equity awareness. Utilizing these metrics quantifies the capability of each approach to not only optimize land use distribution in alignment with overarching urban development goals but also to balance individual interests, emphasizing social equity in the decision-making process. Thus, we focus on maximizing the Global Reward and simultaneously minimizing the Equity Reward in the experiment.

\subsubsection{Urban metrics}
Beyond the algorithmic metrics, the outcomes proposed by each method are assessed from urban planning perspectives, specifically examining the potential improvement in the quality of life for agents in the region. In \hyperlink{https://www.cambridgeredevelopment.org/kendall-square-3}{the government planning guideline of Kendall Square}, two planning metrics, sustainability and diversity, stand out as crucial determinants of its planning targets.

\textit{Sustainability.} A sustainable urban planning strategy can reach the right balance between environmental considerations and economic development. The ideal approach aims to reduce sprawl environmental impact and promote economically thriving communities. Thus, we employ the preceding density metrics of green space and commercial spaces as measures of the sustainability of land utilization in comparisons. 

\textit{Diversity.} Well-balanced urban ecosystems require a diversity of social, economic, and spatial characteristics to achieve optimal urban performance. Using the aforementioned Shannon-Weaver formula, this metric quantifies both the variety of land uses and their even distribution.

These metrics collectively provide valuable insights into the optimization of land usability for the well-being of their inhabitants and the resilience of the cities themselves.


\section{Result Analysis} 
\subsection{Planning results}



\begin{figure}
    \begin{subfigure}{0.55\linewidth}
        \centering
        \includegraphics[width=0.95\linewidth]{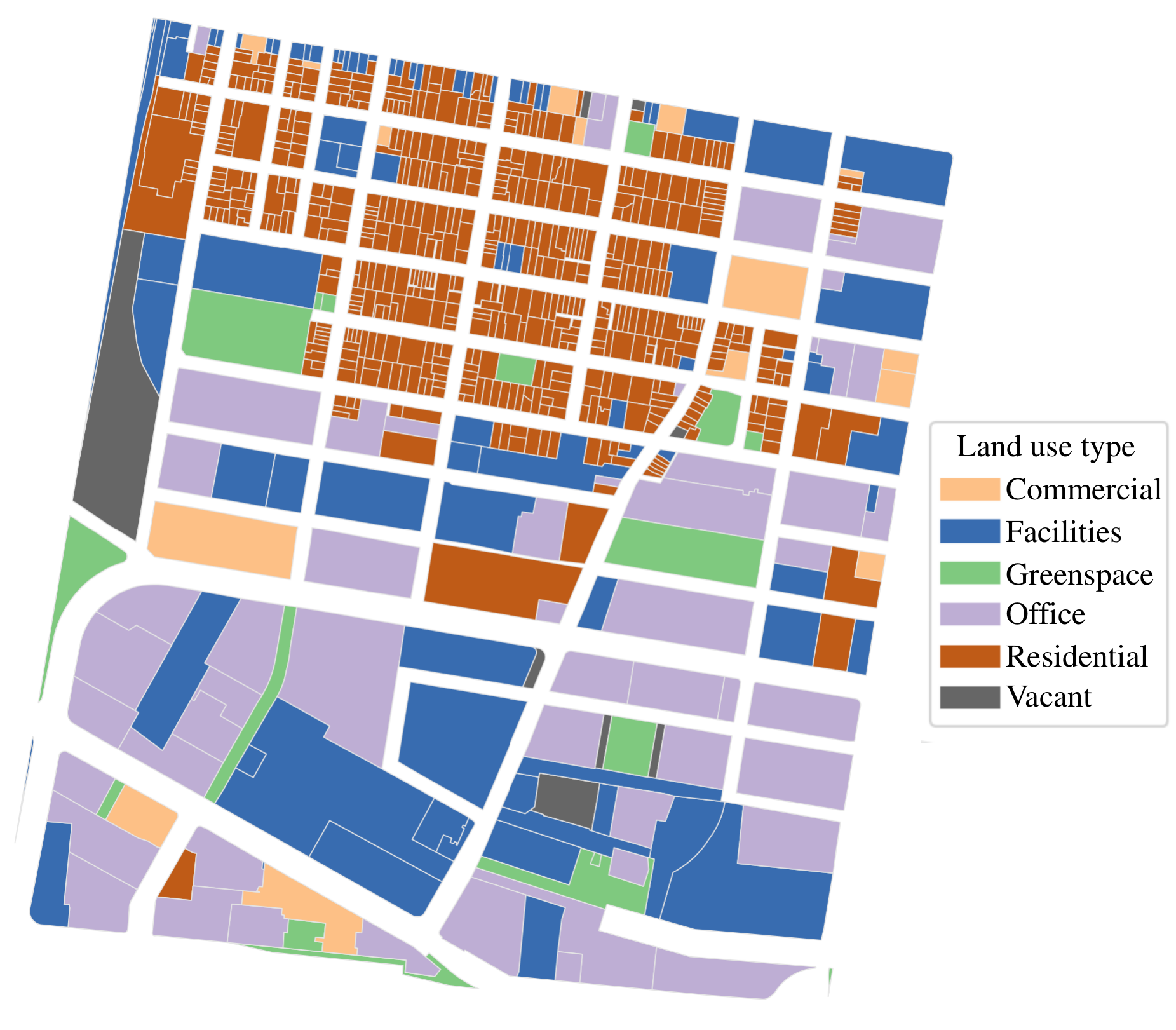}
        \caption{Original status}
    \end{subfigure}%
    \begin{subfigure}{0.45\linewidth}
        \centering
        \includegraphics[width=0.95\linewidth]{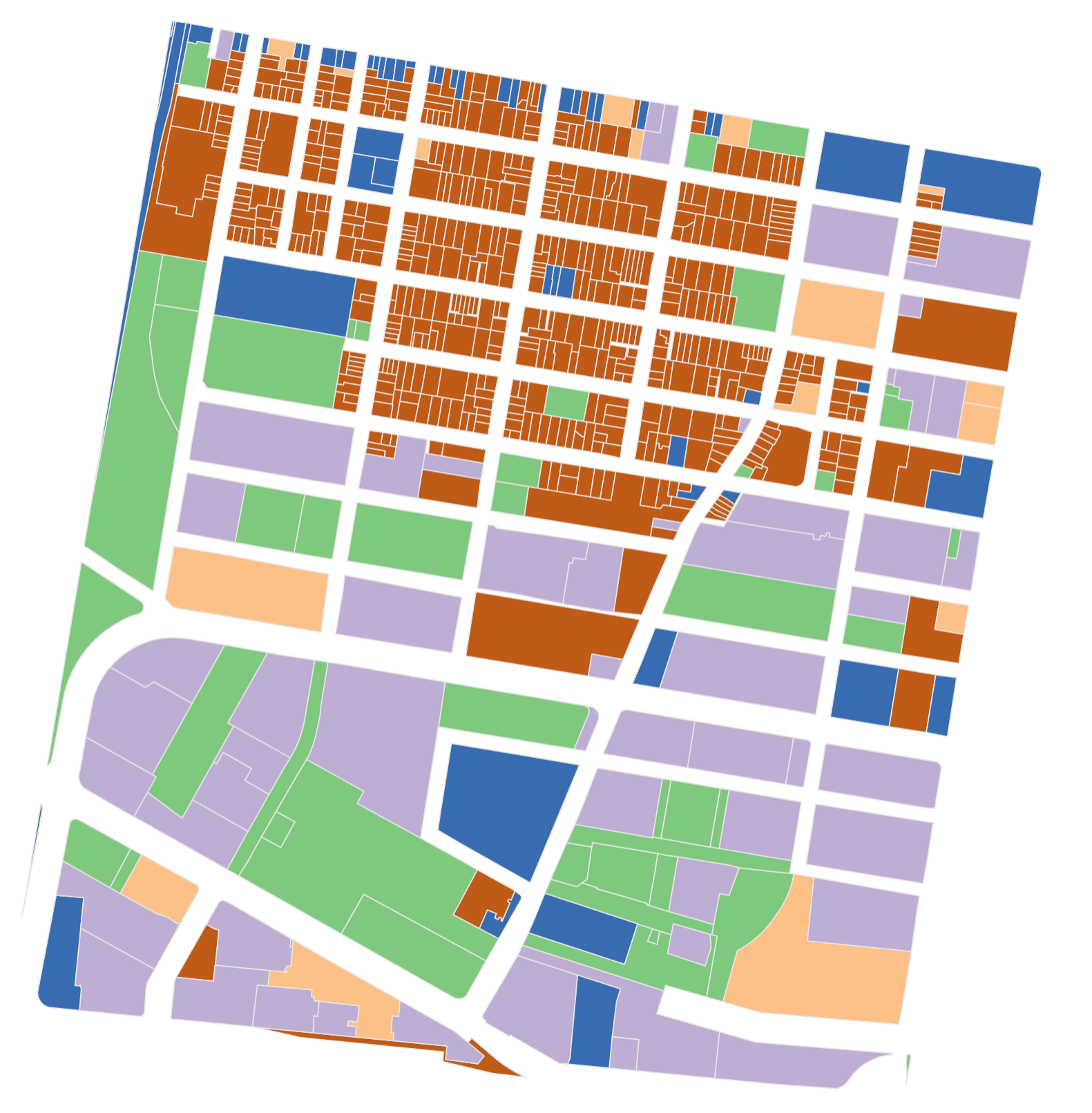}
        \caption{Our planning outcome}
    \end{subfigure}\\
    \caption{Comparisons of our MARL readjustment outcome and original status of land use distribution in Kendall Square.}
    \label{Global_and_participatory}
\end{figure}

The planning outcome generated by consensus-based MARL framework transforms the core area of Kendall Square into a contiguous green space, fostering opportunities for recreation, social gatherings, and environmental enhancement, as shown in Figure \ref {Global_and_participatory}. The vacancies in the northern residential sectors have been repurposed with commercial land uses, promoting residents' easy access to amenities. By reducing office-centric land use and amplifying a diverse range of programs highlighting green spaces and commercials, this approach proposed an idealized urban environment that champions livability.

The result from the top-down approach is relatively conservative but is easier to implement in practice. Meanwhile, the participatory approach provides valuable insights into idealistic alternatives to current planning, envisioning a city model that places greater emphasis on public welfare and sustainability. In conclusion, both decision-making approaches presented planners with a spectrum of possibilities under different values that is meaningful in guiding real-world planning.

\subsection{Comparison analysis}
In the experiment, we benchmarked our proposed approach against three established methodologies: the random method, the greedy algorithm, and DRL. This comparison is conducted within both top-down and participatory planning scenarios across four metrics: Global reward, Equity reward, Sustainability, and Diversity.

\begin{figure}
    \begin{subfigure}{0.5\linewidth}
        \centering
        \includegraphics[width=0.95\linewidth]{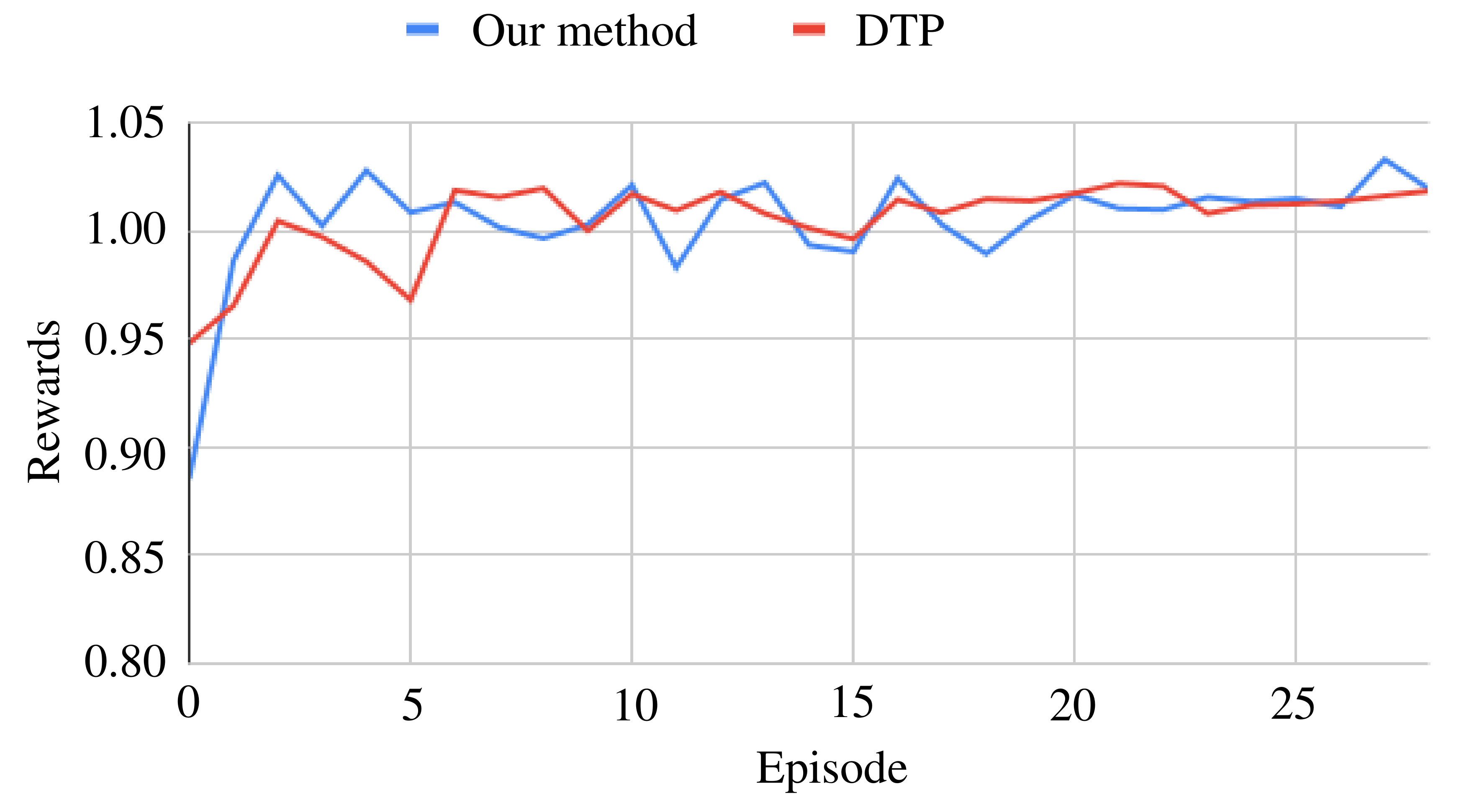}
        \caption{Global reward}
    \end{subfigure}%
    \begin{subfigure}{0.5\linewidth}
        \centering
        \includegraphics[width=0.95\linewidth]{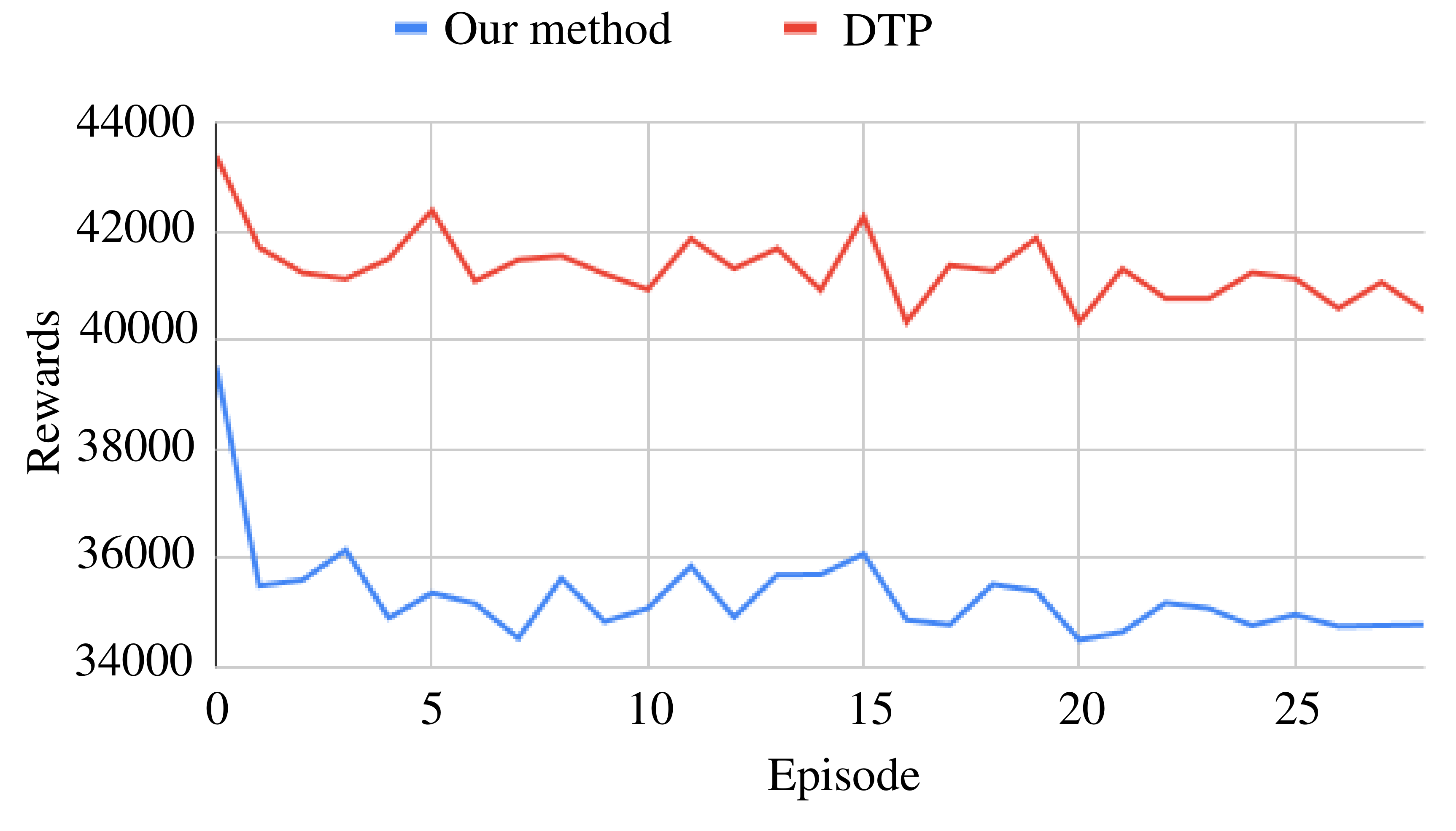}
        \caption{Equity reward}
    \end{subfigure}\\
    \caption{(a) Convergence comparison of Global rewards in MARL method and DTP method. The X-axis is number of episodes. (b) Convergence comparison of Equity rewards in MARL method and DTP method. The X-axis is number of episodes.}
    \label{fig:comparison}
\end{figure}

In the learning process, Figure \ref{fig:comparison} (a) illustrates consensus-based MARL initially starts with a Global reward of 0.8853 and exhibits fluctuations before stabilizing at episode 30 with values of 1.019, which is in proximity to the 0.018 value observed in the DTP method. With respect to equity rewards in Figure \ref{fig:comparison} (b), it shows that the MARL based on participatory planning, outperform deep RL method rooted in top-down planning. It accentuates the potential of participatory-based approaches, particularly the consensus-based MARL, in fostering equity in urban planning decisions. Furthermore, a significant feature of the proposed method is its adaptive capability in dynamic environment, which is reflected in convergent reward increase and leads to achieve a competitive performance with respect to other methods. As further evidence, Table \ref{tab:algorithm} indicates that participatory planning results in the highest global urban performance (0.019) and significant improvements in decision-making equity (34774.515).



\begin{table}[!ht]
\caption{Algorithmic effectiveness comparisons}
\label{tab:algorithm}
\begin{tabular}{ccc}
\bottomrule
Methods  & Global reward  & Equity reward              \\
\hline
RTP & 0.958 & 36084.443                \\ 
RPP & 0.942 & 38697.276                \\ 
GTP & 0.965 & 42117.190       \\
GPP & 0.962 & 35328.027      \\ 
DTP & 1.018 & 40532.082         \\ 
Our method & \textbf{1.019} & \textbf{34774.515}        \\ 
\toprule
\end{tabular}
\end{table}







Beyond evaluating from a reward-centric viewpoint, our planning outcomes significantly outperforms those of alternative approaches from urban planning perspectives. In Table \ref{tab:urban}, it demonstrate the top scores in both sustainability and diversity at 0.299 and 0.449, respectively. Additionally, the readjustment strategy suggested by the consensus-based MARL enhances sustainability and diversity in Kendall Square, observing a growth of 167\% and 8\%. This result also proves the superiority of our method to consistently generate a more diverse readjustment strategy. The limitation of top-down planning approach is emphasized, which overlooks the inclusive and varied insights into land use changes.

\begin{table}[!ht]
\caption{Urban metrics comparisons}
\label{tab:urban}
\begin{tabular}{cccc}
\bottomrule
Methods  & Sustainability & Diversity  \\
\hline
Original Status & 0.112 & 0.416 \\
RTP & 0.191 & 0.401 \\
RPP & 0.155 & 0.406        \\
GTP & 0.170 & 0.404          \\
GPP & 0.168 & 0.404             \\ 
DTP & 0.112 & 0.419         \\ 
Our method & \textbf{0.299} & \textbf{0.449}       \\ 
\toprule
\end{tabular}
\end{table}

\section{Conclusion and future work}
In this study, we first present a spatial graph representing the urban landscape, grounded in urban system theory. Utilizing this data structure, we frame the land use readjustment challenge as a Markov decision process for participatory urban planning. Within this context, we develop a multi-agent reinforcement learning framework containing four tiered metrics based on the individual recognition of land use preferences and urban performance, aiming to produce an optimal land use readjustment blueprint that balances stakeholders' varying interests and fosters sustainable urban development. Through comprehensive experiments focused on a case study in Kendall Square, our innovative approach successfully provides an evidence that participatory planning can enhance the optimization of readjustment strategies through equitable collective decision-making. The results underscore notable advancements both in algorithmic performance and in aligning urban readjustment strategies with government directives in real-world scenarios. Furthermore, the consensus-based MARL model not only elevates urban livability but also promotes equity in participatory urban planning.

In future work, we intend to explore advanced RL techniques and refine reward function strategies to further enhance land use planning and fine-tune the harmonization of stakeholder preferences. Additionally, we plan to integrate this framework into a collaborative platform, drawing upon the expertise of urban professionals, to devise more effective urban planning strategies. 





\section{Citations and References}



\begin{acks}
We sincerely appreciate Yan Zhang from Massachusetts Institute of Technology, and Dr. Yang Liu from Tongji University for their support on this work.
\end{acks}



\bibliographystyle{ACM-Reference-Format} 


\end{document}